\documentclass[nohyperref]{article}

\usepackage{amsthm}

\usepackage{microtype}
\usepackage{amsmath}
\usepackage{amsfonts}
\usepackage{amssymb}
\usepackage{tabu}
\usepackage{url}
\usepackage{graphicx}
\usepackage{booktabs}

\usepackage{xspace}
\usepackage{multirow}
\usepackage{enumitem}
\usepackage{amsmath}
\usepackage{mathtools}

\newcommand{\ms}[1]{\ensuremath{\mathsf{#1}}}

\newcommand{\Z}{\mathbb{Z}}

\newcommand{\C}{\mathbb{C}}

\newcommand{\calC}{\ensuremath{\mathcal{C}}}

\newcommand{\calL}{\ensuremath{\mathcal{L}}}
\newcommand{\calM}{\ensuremath{\mathcal{M}}}

\newcommand{\boldW}{\ensuremath{\mathbf{W}}}
\newcommand{\boldX}{\ensuremath{\mathbf{X}}}
\newcommand{\boldY}{\ensuremath{\mathbf{Y}}}

\newcommand{\ceil}[1]{\left\lceil #1 \right\rceil}

\newcommand{\KeyGen}{\ms{KeyGen}}
\newcommand{\Encrypt}{\ms{Encrypt}}
\newcommand{\Decrypt}{\ms{Decrypt}}

\newcommand{\SwitchKey}{\ms{SwitchKey}}

\newcommand{\Rotation}{\ms{Rotation}}

\newcommand{\Eval}{\ms{Eval}}

\newcommand{\ct}{\ms{ct}}
\newcommand{\sk}{\ms{sk}}
\newcommand{\pk}{\ms{pk}}

\newcommand{\WX}[2]{\ensuremath{\boldW^{#1} \ast \boldX^{#2}}}

\newcommand{\zv}{\mathbf{z}}

\usepackage{tikz}

\usepackage{microtype}
\usepackage{graphicx}
\usepackage{subfigure}
\usepackage{booktabs} 
\usepackage{diagbox}
\usepackage{tabularx}
\usepackage{multirow}
\usepackage{multicol}
\usepackage{comment}

\usepackage{hyperref}

\usepackage[accepted]{icml2022}

\usepackage{amsmath}
\usepackage{amssymb}
\usepackage{mathtools}
\usepackage{amsthm}

\usepackage[capitalize,noabbrev]{cleveref}

\theoremstyle{plain}

\theoremstyle{definition}

\theoremstyle{remark}

\usepackage[textsize=tiny]{todonotes}

\icmltitlerunning{Artemis: HE-Aware Training for Efficient Privacy-Preserving Machine Learning}

\begin{document}

\twocolumn[
\icmltitle{Artemis: HE-Aware Training for Efficient Privacy-Preserving Machine Learning}



\icmlsetsymbol{equal}{*}

\begin{icmlauthorlist}
\icmlauthor{Yeonsoo Jeon}{ut}
\icmlauthor{Mattan Erez}{ut}
\icmlauthor{Michael Orshansky}{ut}
\end{icmlauthorlist}

\icmlaffiliation{ut}{Department of Electrical and Computer Engineering, University of Texas at Austin, Texas, US}

\icmlcorrespondingauthor{Yeonsoo Jeon}{yeonsoo@utexas.edu}

\icmlkeywords{Machine Learning, ICML}

\vskip 0.3in
]



\printAffiliationsAndNotice{}  

\begin{abstract}
Privacy-Preserving ML (PPML) based on Homomorphic Encryption (HE) is a promising foundational privacy technology. Making it more practical requires lowering its computational cost, especially, in handling modern large deep neural networks. Model compression via pruning is highly effective in conventional plaintext ML but cannot be effectively applied to HE-PPML as is. 

We propose Artemis, a highly effective DNN pruning technique for HE-based inference. We judiciously investigate two HE-aware pruning strategies (positional and diagonal) to reduce the number of $\Rotation$ operations, which dominate compute time in HE convolution. We find that Pareto-optimal solutions are based fully on diagonal pruning. Artemis' benefits come from coupling DNN training, driven by a novel group Lasso regularization objective, with pruning to maximize HE-specific cost reduction (dominated by the $\Rotation$ operations). 
We show that Artemis improves on prior HE-oriented pruning and can achieve a 1.2-6x improvement when targeting modern convolutional models (ResNet18 and ResNet18) across three datasets. 





\end{abstract}

\section{Introduction}
\label{Introduction}

The growing interest in Machine Learning as a Service has raised privacy concerns. One such issue is the potential exposure of raw user data to an untrusted server, particularly if it contains personal information. Homomorphic Encryption (HE) has been proposed as a solution to this privacy challenge. HE encrypts the client's data, allowing computations to be performed on ciphertext without allowing the server to learn any information. Indeed, \cite{cryptonets, cryptodl, lola} use Leveled Homomorphic Encryption (LHE) to achieve privacy-preserving machine learning systems using HE.

Despite its potential, privacy-preserving machine learning using LHE is still far from practical due to its high computational complexity. This is because an LHE implementation requires a far larger number of operations compared to computations on plaintext. This is a significant hindrance to using advanced ML methods based on deep neural networks (DNNs) with LHE. Analysis shows that the main cost during inference is due to a large number of one specific HE operation, which performs a rotation in the HE representation of the data~\cite{hunter}. 

Prior research has suggested that the number of $\Rotation$ operations can be reduced in inference via model pruning of specific weight patterns~\cite{hunter}. There are two such patterns that reflect the computational properties of the convolution operators mapped to the HE domain.  
The first pattern, which we refer to as \emph{positional pruning}, prunes weights in the same position of multiple channels in a filter. The second method, which we call \emph{diagonal pruning}, operates at the level of entire channels, pruning entire channels lying on a diagonal of a flattened kernel. 

In this work, we develop a method for pruning that significantly outperforms prior work on HE-specific pruning. We achieve this by showing how training can be modified to better prune CNN models for reducing the number of $\Rotation$, and hence overall HE computational cost. Specifically, we show how \emph{group Lasso regularization} can be used to impose structure on the weights of the model such that HE-specific pruning is much more effective. 

We call our approach \emph{Artemis} and evaluate it on a modern CNN (ResNet18 and ResNet50) with three distinct image datasets. We analyze the trade-off between the number of $\Rotation$ operations and accuracy. 
We find that Pareto-optimal solutions do not, in fact, utilize any positional pruning, and are entirely dominated by points produced by diagonal pruning. Thus, using only diagonal regularization and diagonal pruning is sufficient for finding Pareto optimal solutions. 
On three datasets, our algorithm achieves a 1.2-6X greater cost reduction compared to the state-of-the-art method at equal accuracy~\cite{hunter}.

In summary, we make the following main contributions:
\begin{itemize}
\item We perform a detailed analysis of HE inference costs to demonstrate that the number of $\Rotation$ operations dominates run time.
\item We demonstration that conventional structured pruning is not helpful in HE, and investigate two model pruning strategies (positional and diagonal) to reduce the number of $\Rotation$ operations in HE convolution.
\item We make the important observation that some deep networks are not compressible (prunable) under HE-required structure, while still being compressible for plaintext inference.
\item We develop a highly effective DNN training flow based on HE-aware Lasso regularization that results in models that required 1.2-6X fewer rotations compared to the exiting state of the art approaches.
\end{itemize}

\section{Background}
\label{Background}

\subsection{Homomorphic Encryption}
Homomorphic Encryption (HE) is an encryption scheme that permits computation over the ciphertext without decrypting it. Its homomorphic property is due to that the decryption algorithm preserves the group structure of the ciphertext space. Fully Homomorphic Encryption (FHE), which allows arbitrary computation over the ciphertext, has been a challenge until recently. Since Gentry first introduced FHE \cite{gentry}, several FHE schemes, including \cite{bfv, bgv, ckks, tfhe}, have been proposed.

Our paper focuses on HE schemes that in general support SIMD operations that include \cite{bfv, bgv}, because they are most efficient. However, for the implementation, this paper chooses CKKS \cite{ckks}.

The CKKS scheme consists of five algorithms: $\KeyGen$, $\Encrypt$, $\Decrypt$, $\Eval$, and $\SwitchKey$ ($\Rotation$). The plaintext space $\calM = R = \Z[X]/(X^N + 1)$, the cipertext space $\calC = R_q^2$ where $R_q = R/qR$, and secret key is $\sk = (1, s) \in R^2$. $\KeyGen$, $\Encrypt$, and $\Decrypt$ are similar to those of other lattice-based public key encryption schemes.

$\Eval$ computes addition and multiplication over the ciphertext. Given two plaintext, $m_1, m_2 \in \calM$, and arbitrary function $f$, $\Eval(\pk, \ct_1, \ct_2, f) \rightarrow \ct_f$ such that $\Decrypt(\sk, \ct_f) = f(m_1, m_2)$, where $\ct_i$ is an encryption of $m_i$. CKKS supports addition and multiplication over the ciphertext. 


$\SwitchKey$ switches the underlying key of a ciphertext. For example, $\SwitchKey(\ms{swk}_{s \rightarrow k}, \ct_s) \rightarrow \ct_k$ where $\ms{swk}_{s \rightarrow k}$ is a public key for switching the key from $s$ to $k$. $\ct_s$ is a ciphertext under secret key $s$ and $\ct_k$ is a ciphertext under secret key $k$. 

Importantly, $\SwitchKey$ is required in performing a permutation (or a rotation) of a message. CKKS distinguishes the terms message and plaintext. The message space is $\C^{N/2}$. $\ms{Encode}, \ms{Decode}$ algorithms are canonical embedding between $\C^{N/2}$ and $\calM$ (and inverse to each other). To decode a plaintext $m(X) \in \calM$, evaluate the $2N$-th primitive roots of unity, namely, $\ms{Decode}(m) \rightarrow \zv$, where $\zv = (m(\zeta^{5^0}), \cdots, m(\zeta^{5^{N/2-1}}))$. 

Rotation works as follows. Let $\ct_s = (c_0, c_1)$. Consider $\ct_{s'}(X^{5^r}) = (c_0(X^{5^r}), c_1(X^{5^r}))$, where $s' = s(X^{5^r})$. Then, $\ct_{s'}(X^{5^r})$ is an encryption of $m(X^{5^r})$, which is a plaintext of rotate messages. $\ms{Decode}(m(X^{5^r})) \rightarrow \zv' \in \C^{N/2}$, where $\zv'_i = \zv_{r+i}$. Therefore, the rotation reduces to $\SwitchKey(\ms{swk}_{s \rightarrow s'}, \ct_s) \rightarrow \ct_{s'}$, where $s' = s(X^{5^r})$. Hereinafter, because $\Rotation$ is implemented by $\SwitchKey$, we refer to the operation as $\Rotation$.

The operations of HE schemes can be viewed as SIMD operations. The positions of the elements of a message $\zv \in \C^{N/2}$ are called `slots'. In the $\Eval$ algorithm, element-wise addition and multiplication are performed, similar to SIMD operations. To perform addition between slots, the ciphertext must be rotated to align the slots.



\subsection{ML Model Compression Techniques}
Model compression is well-established in the area of conventional DNNs. In \cite{unstructuedPruning}, the authors note that many CNN models contain redundant parameters and that removing some of these parameters does not significantly reduce model accuracy. This method is effective in reducing model size and reducing the number of MACs required. In \cite{unstructuedPruning}, pruning is done in an unstructured way. Unstructured pruning involves identifying and removing less important (small-magnitude) parameters without adhering to any specific pattern or structure. 

Unstructured pruning, however, is not easily exploitable in practical accelerators. In \cite{lss, granularitySparsity}, the authors introduce the idea of structured pruning. Their strategy involves grouping the weights in a model in various ways, including row-wise, channel-wise, or filter-wise. Structured pruning results in models that are more hardware-friendly, as they exhibit coarse-grained sparsity that can be more easily leveraged by specialized hardware for faster inference.

It has been found that the training can be done in a way that encourages pruning by employing specific regularization techniques during training. \cite{lss} shows that group Lasso can effectively zero out weights in a group. Consequently, incorporating group Lasso during training helps the structured pruning of weights.

Nonetheless, conventional model compression methods are not effective in reducing computation costs when applied to HE. This is because the messages are encoded and encrypted in multiple ciphertexts and have to be computed only with SIMD operations and the main cost of HE is $\Rotation$ operations. To address this \cite{hunter} introduces new methods for grouping weights within the context of HE that exploit the algebraic structure of HE to reduce the number of $\Rotation$ operations.

\section{HE Structure-Aware Training}

In this section, we propose training the model to improve pruning, which can significantly reduce the computational cost. First, we describe how convolution is done in a HE setting. Then, we illustrate two pruning methods to reduce the number of rotations. Finally, we propose group Lasso regularization and a training method that increases the opportunity to reduce cost while preserving accuracy.

\begin{figure}[t!]
\vskip 0.2in
    \begin{subfigure}
    \centering
    \includegraphics[width=\columnwidth]{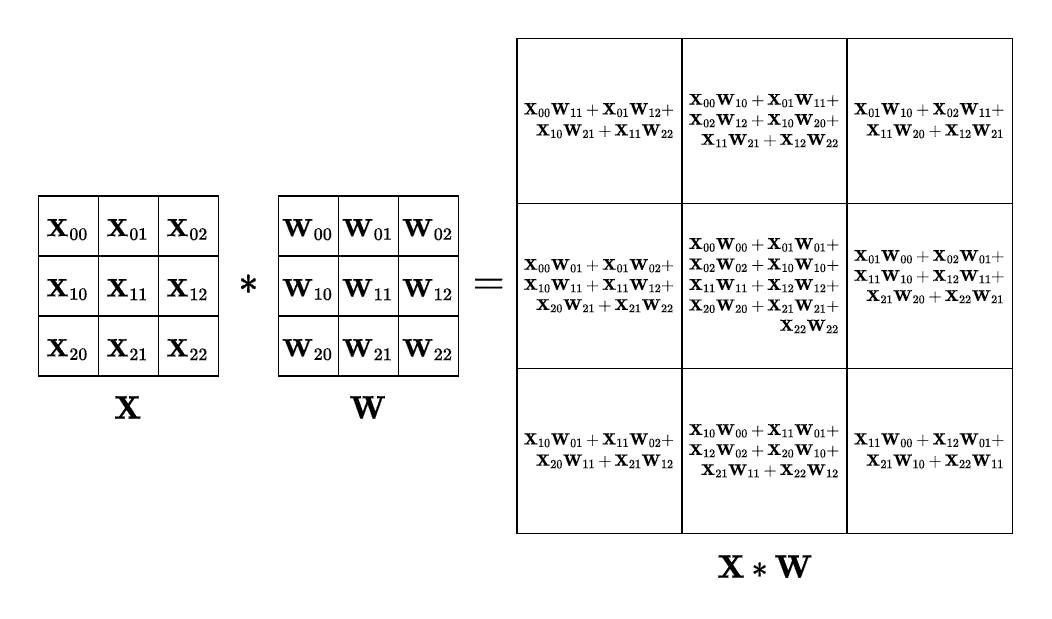}
    \vskip -0.15in
    \caption{Conventioanl 2D convolution $(w_{out}, h_{out}, f) = (3, 3, 3)$.}
    \label{conv}
    \end{subfigure}%
    \vskip 0.2in
    \begin{subfigure}
    \centering
    \includegraphics[width=\columnwidth]{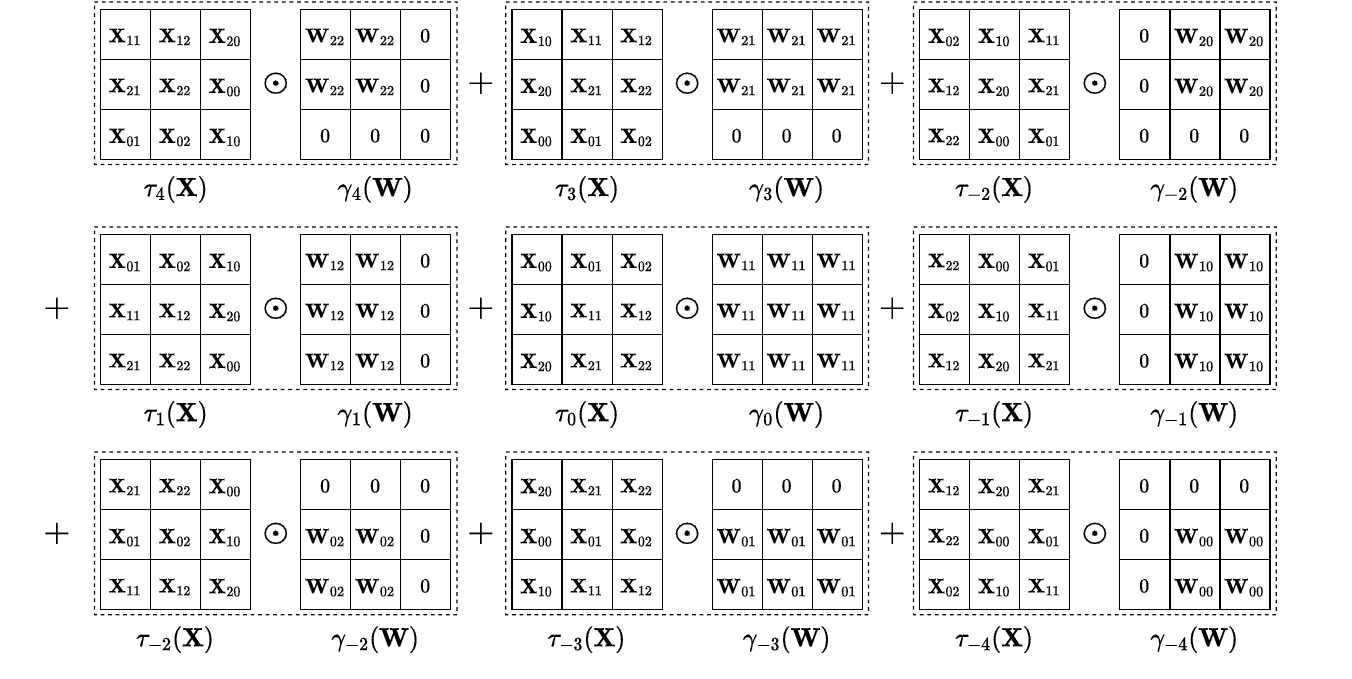}
    \vskip -0.15in
    \caption{SISO convolution with rotating activations. The result is equivalent to the result of \cref{conv}.}
    \label{rconv}        
    \end{subfigure}%
\vskip -0.2in
\end{figure}

\subsection{Convolution with Rotating Activation}
\label{Convolution with Rotating Activation}
SIMD-based HE schemes, including CKKS, the ciphertext requires that be rotated to allow slot-based accumulation. This makes HE different compared to the plain case. Below is the convolution with rotating activations introduced in \cite{gazelle} and improved in \cite{cheetah}. Recall that $R = \Z[X]/(X^N + 1)$ and there are $N/2$ SIMD slots in a ciphertext. Typically, the length of an activation vector increases during inferences, and thus, the activation has to be encrypted into multiple ciphertexts. Let $c_n$ be the number of channels packed into a single ciphertext. Then, $c_n = \ceil{\frac{N/2}{w_0 h_0}}$, where $w_0, h_0$ are the width and height of the input channel. 

In PPML frameworks including \cite{gazelle, cheetah, hunter}, convolution is divided into two sub-operations, element-wise product and rotate-and-sum (RaS). Conventional plaintext convolution uses a sliding window dot-product. \cite{gazelle} shows how to do convolution with SIMD-based HE by rearranging the channel weights and rotating the input activation. They introduce the convolution method of Single Input Single Output channel convolution (SISO) and generalize it into Multiple Input Multiple Output channel convolution (MIMO). Consider an input activation of $(w_{in}, h_{in}) = (3, 3)$ and a filter of $(w_{out}, h_{out}, f) = (3, 3, 3)$, where $w, h$ are the width, the height and $f$ be the size of a filter, respectively. Let $\boldX$ be (encryption of) the activation vector and $\boldW$ be the weight. Let $\boldX$ and $\boldW$ are $3\times 3$ matrices. We write $\boldX_{ij}$ to denote the $i$-th row and $j$-th column of $\boldX$. 

We focus on two elements $(\boldX \ast \boldW)_{00}$ and $(\boldX \ast \boldW)_{11}$, which are the element $(00)$ and $(11)$ of the convolution result. By the definition of convolution, $(\boldX \ast \boldW)_{00} = \boldX_{00}\boldW_{11} + \cdots + \boldX_{11}\boldW_{22}$ and $(\boldX \ast \boldW)_{11} = \boldX_{00}\boldW_{00} + \cdots + \boldX_{11}\boldW_{11} + \cdots + \boldX_{22}\boldW_{22}$. Note that both elements have terms containing weight $\boldW_{11}$, and are multiplied by the activation at the same position of the activation, $\boldX_{00}$. The same case holds for other positions. Therefore, we use $\boldW_{11}$ to construct an arranged weight $\gamma_0(\boldW)$, a $3 \times 3$ matrix where all elements are $\boldW_{11}$. Then, we compute the partial multiplication by element-wise multiplying $\boldX \odot \gamma_0(\boldW)$.

We now consider the term $\boldX_{12}\boldW_{12}$. Note that the term is also added to compute $(\boldX \ast \boldW)_{11} = \boldX_{00}\boldW_{00} + \cdots + \boldX_{11}\boldW_{11} + \boldX_{12}\boldW_{12} + \cdots + \boldX_{22}\boldW_{22}$. Finally, at this moment, we rotate the activation. Let $\tau_i(\boldX)$ be the left rotation of $\boldX$ by $i$, namely $(\tau_1(\boldX))_{11} = \boldX_{12}$. We take $\boldW_{12}$ to construct an arranged weight $\gamma_1(\boldW)$, a $3 \times 3$ matrix where the elements of left two columns are $\boldW_{12}$ and the rightmost column are 0. Then, clearly, $(\tau_1(\boldX) \odot \gamma_1(\boldW))_{11} = \boldX_{12}\boldW_{12}$. Moreover, consider $(\tau_1(\boldX) \odot \gamma_1(\boldW))_{00} = \boldX_{01}\boldW_{12}$. This term is a partial multiplication of $(\boldX \ast \boldW)_{00}$. The rotated activation, $\tau_i(\boldX)$ can be reused in computing the partial multiplications at the other positions. Therefore, with 8 rotations without the trivial rotation of 0, multiplying by the weight in the convolution operation can be computed. Summarizing the above method, we have the following equation:

\begin{equation}
   \boldX \ast \boldW = \sum \tau_i(\boldX) \odot \gamma_i(\boldW) 
\end{equation}


\cref{conv} and \cref{rconv} illustrate conventional 2D convolution and its realization with rotating activation vector. \cref{rconv} shows all $\tau_i(\boldX)$ and $\gamma_i(\boldW)$ for all $i$ and shows that the result of rotating the activation and rearranging the weights is equivalent to that of the conventional convolution. 

Hereinafter, we refer to this method $\ms{SISO}$. One can easily extend $\ms{SISO}$ from the above to compute multiple channel 2D-convolutions in a ciphertext simultaneously, or $\ms{MIMO}$. Consider the convolution of $(c_{in}, c_{out}, f) = (4, 4, 3)$ with $c_n = 4$. (For simplicity, we choose $c_n = c_{in} = c_{out}$ for the analysis.) Let $\boldX^{i}$ to denote the $i$-th channel of the activation and $\boldW^{ij}$ to denote the $j$-th channel of $i$-th filter i.e. these are the 2D sub-matrices of $\boldX$ and $\boldW$, respectively. 

\begin{figure}[t!]
\vskip 0.2in
\begin{subfigure}
    \centering
    \includegraphics[width=\columnwidth]{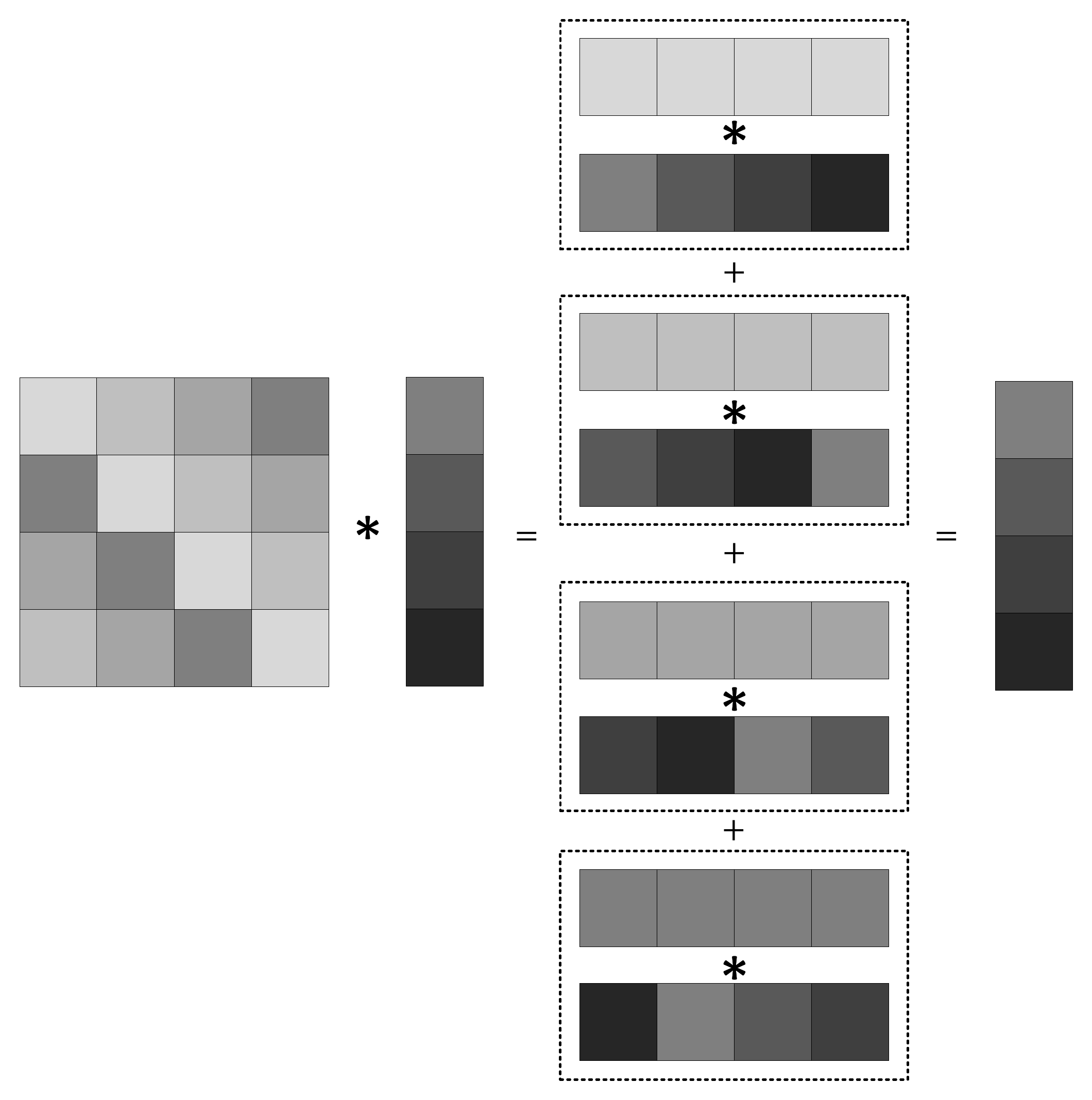}
    \vskip -0.2in
    \caption{Squares with light gray are activations and squares with dark gray are weights. Multiplying with weights is realized with $\ms{SISO}$. The figure shows a method of rotating weights to add partial results.}
    \label{diagonaladd}
\end{subfigure}%
    \vskip -0.1in
\begin{subfigure}
    \centering
    \centerline{
    \includegraphics[width=0.75\columnwidth]{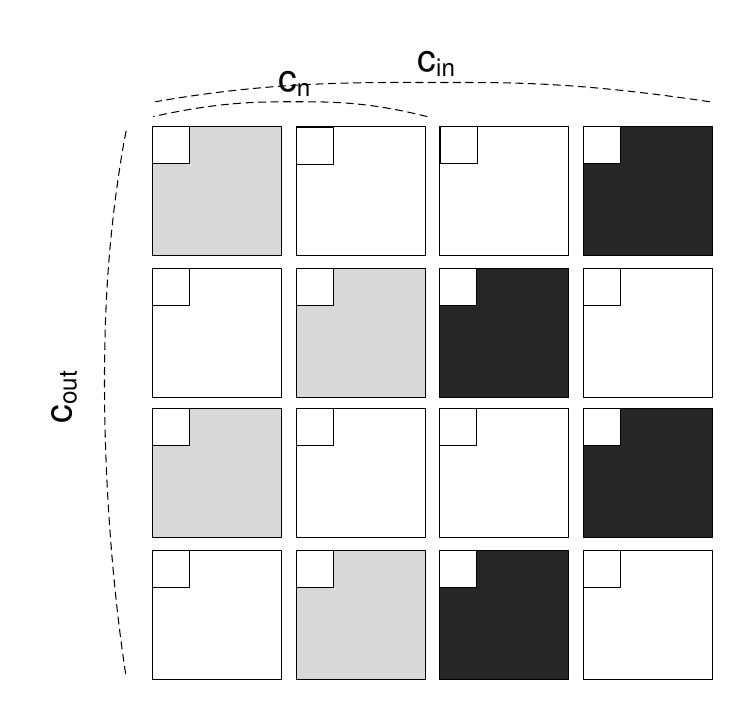}}
    \vskip -0.25in
    \caption{Two pruning methods. (a) Position pruning to reduce the number of $\tau_i(\boldX)$ operations; (b) Diagonal pruning to reduce the number of $\pi_i(\boldX)$ operations. The pruned weights are shown as white squares.}
    \vskip -0.1in
    \label{pruning}        
\end{subfigure}%
\end{figure}

By definition of the convolution, we compute the output as follows:

\begin{equation}
\resizebox{0.9\columnwidth}{!}{%
$
    \begin{bmatrix}
        \boldY^{0} \\
        \boldY^{1} \\
        \boldY^{2} \\
        \boldY^{3}
    \end{bmatrix}
    =
    \begin{bmatrix}
        \WX{00}{0} + \WX{01}{1} + \WX{02}{2} + \WX{03}{3}\\
        \WX{10}{0} + \WX{11}{1} + \WX{12}{2} + \WX{13}{3}\\
        \WX{20}{0} + \WX{21}{1} + \WX{22}{2} + \WX{23}{3}\\
        \WX{30}{0} + \WX{31}{1} + \WX{32}{2} + \WX{33}{3}
    \end{bmatrix}
$
}
\end{equation}

where $\WX{ij}{j}$ will be in fact computed as $\ms{SISO}(\boldW^{ij}, \boldX^{j})$. We can rearrange the equations as follows:
\begin{equation}
\resizebox{0.9\columnwidth}{!}{%
$
    \begin{bmatrix}
        \boldY^{0} \\
        \boldY^{1} \\
        \boldY^{2} \\
        \boldY^{3}
    \end{bmatrix}
        =     
    \begin{bmatrix}
        \WX{00}{0} \\
        \WX{11}{1} \\
        \WX{22}{2} \\
        \WX{33}{3} 
    \end{bmatrix}
    +
    \begin{bmatrix}
        \WX{01}{1} \\
        \WX{12}{2} \\
        \WX{23}{3} \\
        \WX{30}{0} 
    \end{bmatrix}
    +
    \begin{bmatrix}
        \WX{02}{2} \\
        \WX{13}{3} \\
        \WX{20}{0} \\
        \WX{31}{1} 
    \end{bmatrix}
    +
    \begin{bmatrix}
        \WX{03}{3} \\
        \WX{10}{0} \\
        \WX{21}{1} \\
        \WX{32}{2} 
    \end{bmatrix}
$
}
\label{rotateAdd}
\end{equation}

Note that the \cref{rotateAdd} consists of permutation of $\boldX$. Therefore, we exploit the structure to add partial results. \cref{diagonaladd} illustrates the method. 

Let $\pi_j(\boldX)$ be the the left rotation of $\boldX$ by $j$ channels. Also, let $\delta_{u}(\boldW)$ be the $u$-th diagonal of $\boldW$, $\delta_u(\boldW) = [\boldW^{(u)0} | \boldW^{(u+1)1} | \boldW^{(u+2)2} | \boldW^{(u+3)3}]$. Then, \cref{rotateAdd} is equivalent to 

\begin{equation}
    \boldY = \sum \pi_j(\ms{SISO}(\delta_{j}(\boldW), \boldX))
\end{equation}

We can now compute the number of $\Rotation$ operations for the convolution required by the above-described algorithm. $(f^2 - 1)$ $\Rotation$ operations are required for each $\ms{SISO}$, and $(c_{n} - 1)$ $\Rotation$ operations are required for RaS excluding the trivial rotations ($\tau_0(\boldX)$ and $\pi_0(\boldX)$). Therefore, the total number of $\Rotation$ operations is $(f^2-1) + (c_{n} - 1)$. More generally, the number of $\Rotation$ operations required in a convolution layer of dimensions $c_{in}, c_{out}$ and filter size of $f^2$ is:

\begin{equation}
N_{\ms{Rotation}} = \left((f^2 - 1) + \frac{(c_n - 1)c_{out}}{c_n} \right) \frac{c_{in}}{c_n}    
\end{equation}


    


\subsection{Pruning for HE Efficiency}
\label{Pruning for HE Efficiency}
There are two possible ways to reduce the number of $\Rotation$ operations via pruning. The first possibility for pruning, \textit{position pruning} is to reduce the $\Rotation$ operations required for computing $\ms{SISO}$. When performing $\ms{SISO}$, the weight is restructured and the activation is rotated. If all weights in the same position, across all the channels, are eliminated, resulting in $\boldW^{ij}_{uv} = 0$ for some $u, v$ for all $i, j$, allows dropping a $\ms{Rotation}$. The decision on whether to prune is made with respect to a threshold which we take to be $L_2$ norm of the group of coefficients to be removed. The group of weights is pruned out if $L_2\left(\boldW_{uv}\right) = \sqrt{\sum_{i, j}\left(\boldW_{uv}^{ij}\right)^2} $ is less than a threshold. 

Another way to decrease the number of $\Rotation$ operations is by reducing the rotations needed to add partial results. For example, we see that if $[\boldW^{01} | \boldW^{12} | \boldW^{23} | \boldW^{30}] = 0$ in \cref{rotateAdd}, one can avoid a corresponding $\ms{Rotation}$ operation. This second pruning method, which we call \textit{diagonal pruning}, reduces the $\Rotation$ operations required for RaS. It requires that all the weights in the same diagonals be eliminated concurrently so that $\delta_j(\boldW) = 0$ for some $j$. Similarly to position pruning, we zero out the group of weights if $L_2(\delta_j\left(\boldW\right)) = \sqrt{\sum_{u,v}\left(\delta_j\left(\boldW \right)_{uv} \right)^2}$ is less than a threshold.

The two pruning strategies are illustrated in \cref{pruning}. The figure shows the case of $(c_{in}, c_{out}, c_n) = (4, 4, 2)$, where two channels are packed in a ciphertext. The white squares show the pruned weights. 

The number of $\ms{Rotation}$ operations reduced by the proposed method can be estimated as follows. Let $\alpha$ denote the ratio of non-zero weight after position pruning. As a result of this pruning, the number of $\Rotation$ in $\ms{SISO}$ decreases in proportion to $\alpha$. Similarly, let $\beta$ be the ratio of non-zero weight after diagonal pruning. This pruning reduces the number of $\Rotation$ operations involved in RaS in proportion to $\beta$. Consequently, the number of $\Rotation$ operations is now reduced to
\begin{equation}
N_{\ms{Rotation}} = \left(\alpha (f^2 - 1) + \beta \frac{(c_n - 1)c_{out}}{c_n}\right)\frac{c_{in}}{c_n}
\label{cost}
\end{equation}

\subsection{Training for HE-Aware Structured Pruning}
\label{Training for Pruning}
The core innovation of our work is the development of an effective model-training strategy that results in group structure needed for aggressive pruning. As observed above, pruning can only be effective if opportunities for diagonal and position-base pruning abound. We demonstrate how group Lasso regularization can be leveraged to produce such structure. 
Regularization, most commonly $L_2$-norm or Ridge-regularization, is widely used in deep learning to prevent overfitting \cite{deeplearningbook}. In group Lasso \cite{grouplasso}, the regularization term can be viewed as a mixture of the $L_1$- and $L_2$-type penalty. The result is that it trains a model in a way that forces an entire group of weights to be minimized together.

Both diagonal and position-based pruning can be enhanced via structured regularization, leading us to introduce two group Lasso terms.
To regularize the weights for position pruning for a given layer $\ell\in L$, we define position regularization $R_{g, p}\left(\boldW^{(\ell)}\right) = \sum_{u, v} \left(\sqrt{\sum_{i, j}\left(\left(\boldW^{(\ell)}\right)_{uv}^{ij}\right)^2} \right)$, where $u, v$ are the rows and column of the weights.
The regularization term is evaluated for each layer. 

To regularize the weights for diagonal pruning for a given layer $\ell\in L$, we define diagonal regularization $R_{g, d}\left(\boldW^{(\ell)}\right) = \sum_{j} \left( \sqrt{\sum_{u,v}\left(\delta_j\left(\boldW^{(\ell)} \right)_{uv} \right)^2} \right)$. This term is also computed for each layer.

The complete loss function used for training is:
\begin{multline}
    \calL(\boldW) = \calL_D(\boldW) + \lambda R(\boldW) + \\ 
    \lambda_p \sum_{\ell=1}^{L} R_{g, p}(\boldW^{(\ell)}) + \lambda_d \sum_{\ell=1}^{L} R_{g, d}(\boldW^{(\ell)})
\label{loss}
\end{multline}
    
where $\calL(\cdot)$ is the loss function, $\calL_D(\cdot)$ is the loss on the data, $R(\cdot)$ is non-structured $L_2$ regularization for all weights, and $R_{g, p}, R_{g, d}$ are position and diagonal regularization, respectively. The regularization factors $\lambda_p$ and $\lambda_d$ are hyper-parameters for the regularization and $\lambda$ is a hyper-parameter for non-structured regularization.

\begin{figure*}[t!]
\begin{multicols}{2}
    
    \centering
    \subfigure[]{\includegraphics[width=\linewidth]{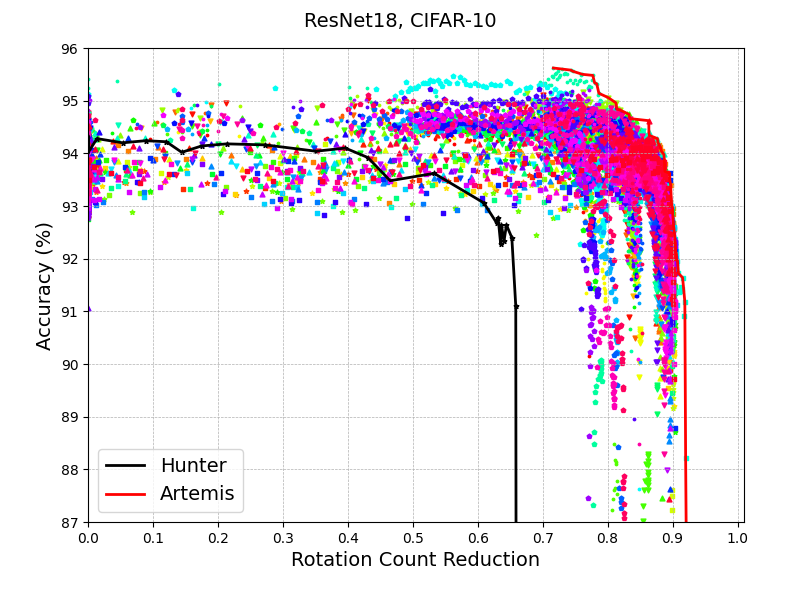}\label{cifar-10 full}} 
    \subfigure[]{\includegraphics[width=\linewidth]{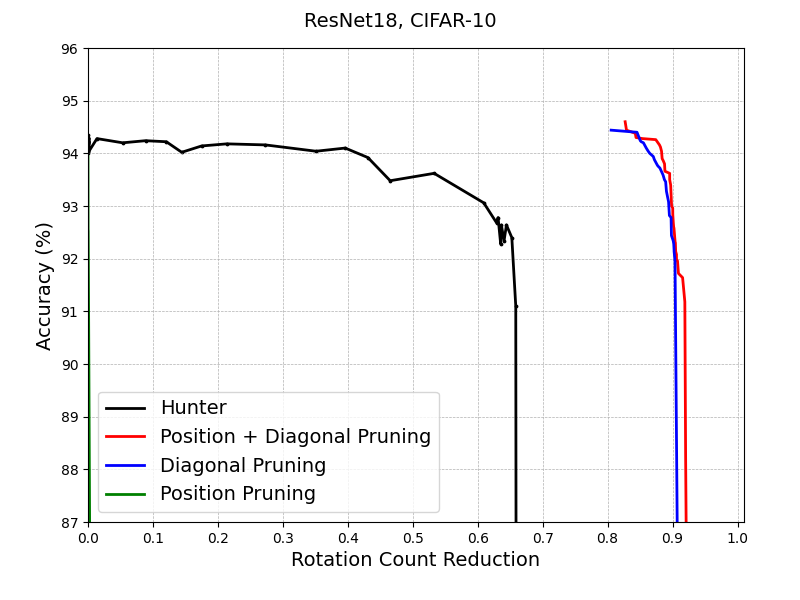}\label{cifar-10 simple}} 
\end{multicols}
\vskip -0.2in
    \caption{(a) A complete set of Artemis solutions (different regularization factors) and the baseline (Hunter); (b) Pareto-optimal Artemis solutions are almost entirely due to diagonal, not positional, pruning.}
\label{scatter}
\end{figure*}

\begin{figure*}[t!]
\begin{multicols}{2}
\centering
    \subfigure[]{\includegraphics[width=\linewidth]{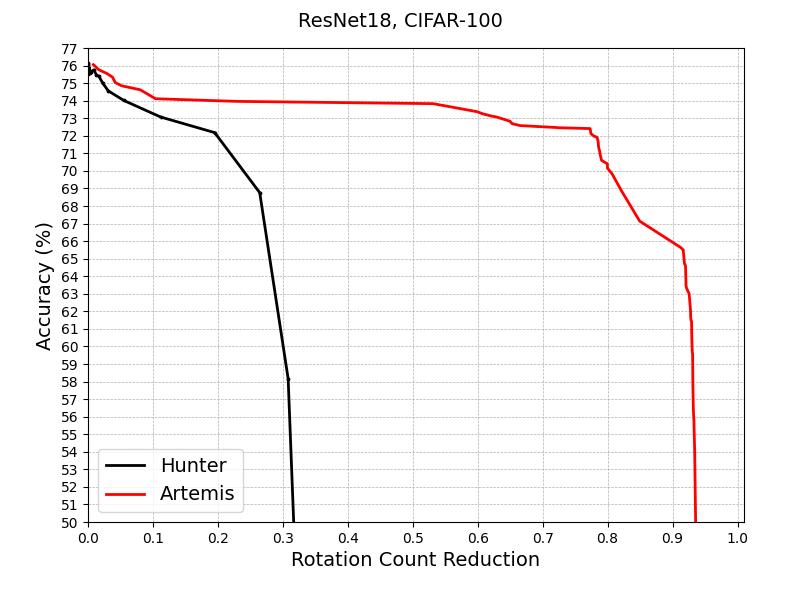} \label{cifar-100}} 
    \subfigure[]{\includegraphics[width=\linewidth]{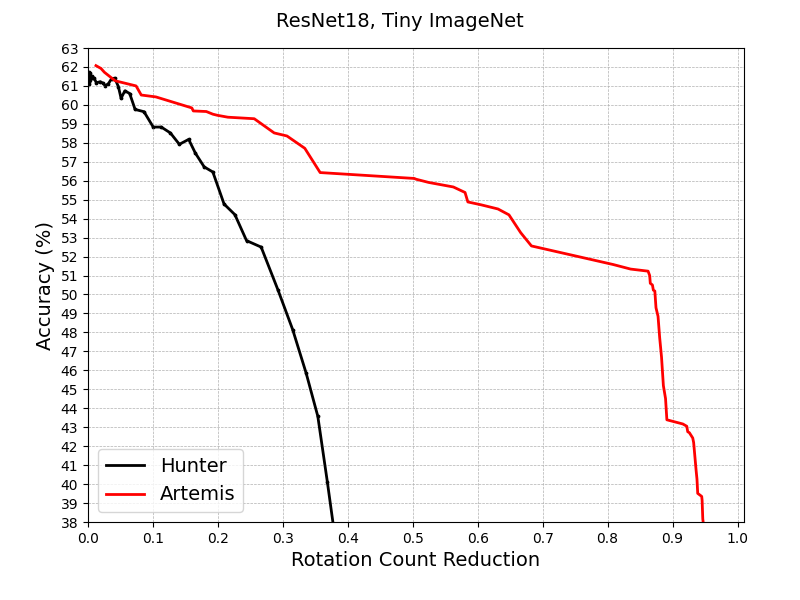} \label{tiny}} 
\end{multicols}
\vskip -0.2in
\caption{Pareto-optimal Artemis solutions (without position pruning) and the baseline for ResNet18 on (a) CIFAR-100 and (b) Tiny ImageNet.}
\label{larger dataset resnet18}
\end{figure*}

\begin{figure*}[t!]
\begin{multicols}{3}
\centering
    \subfigure[]{\includegraphics[width=\linewidth]{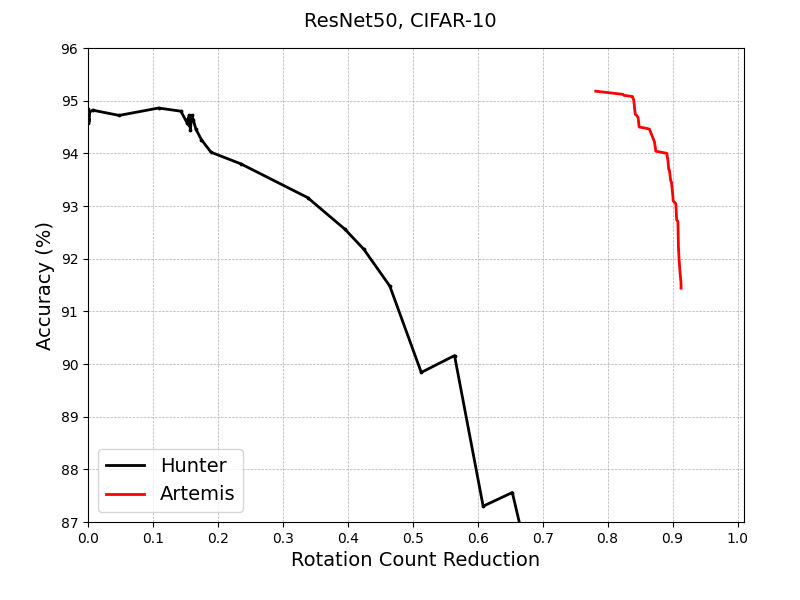} \label{cifar-10resnet50}} 
    \subfigure[]{\includegraphics[width=\linewidth]{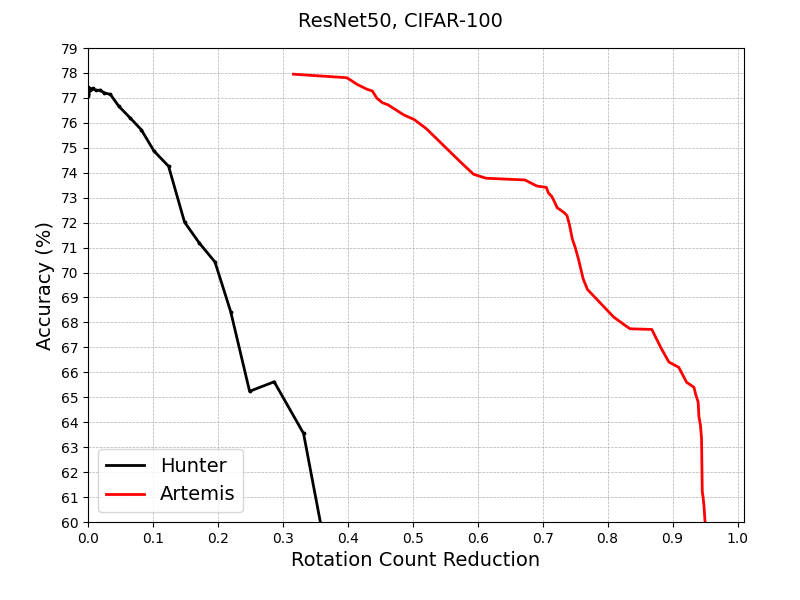} \label{cifar-100resnet50}} 
    \subfigure[]{\includegraphics[width=\linewidth]{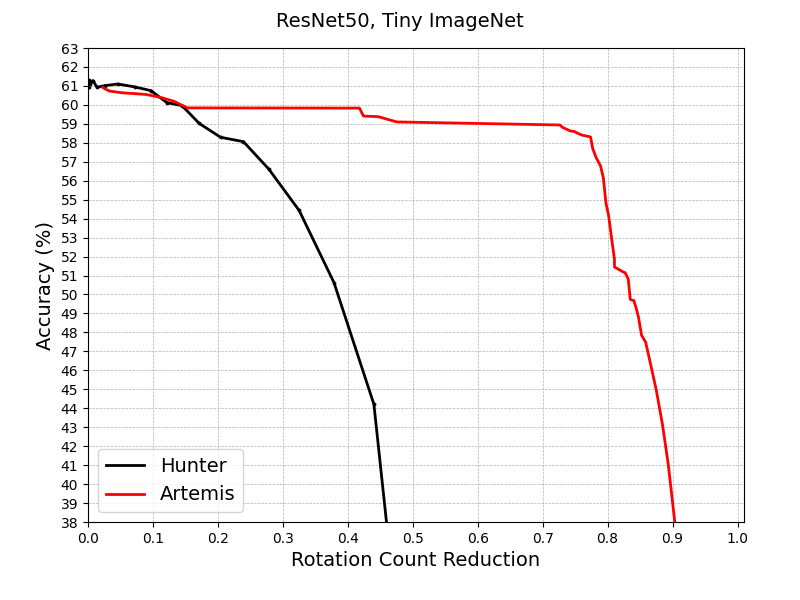} \label{tinyresnet50}} 
\end{multicols}
\vskip -0.2in
\caption{Pareto-optimal Artemis solutions (without position pruning) and the baseline for ResNet50 on (a) CIFAR-100, (b) CIFAR-100, (c) Tiny ImageNet.}
\vskip -0.1in
\label{resnet50}
\end{figure*}

\begin{figure}[t!]
    \centering
    \includegraphics[width=\linewidth]{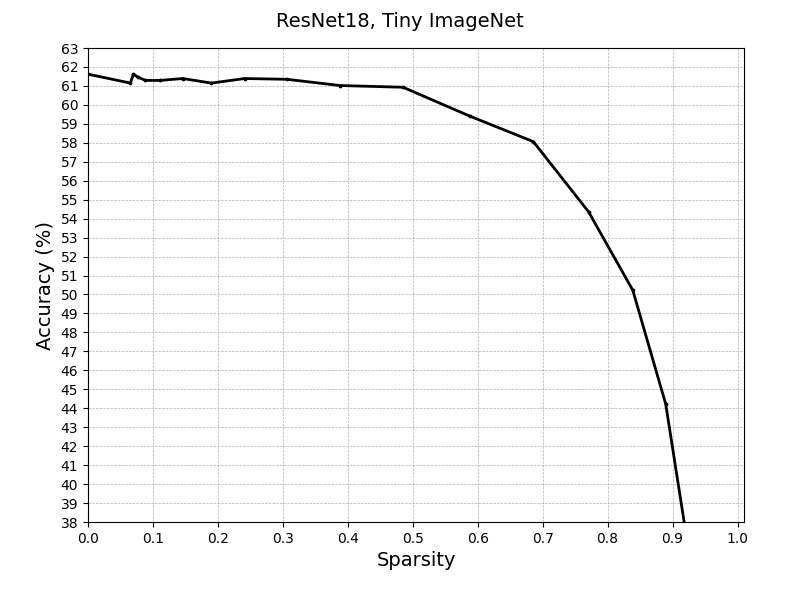}
    \vskip -0.2in
    \caption{Conventional channel pruning on ResNet18/ Tiny ImageNet.}
    \vskip -0.2in
    \label{resnet18 channel prune}
\end{figure}

\section{Results}

We evaluate Artemis using ResNet18 and ResNet50 on the CIFAR-10, CIFAR-100, and Tiny ImageNet datasets. Training uses the Xavier \cite{xavier} method to initialize weights. 
For CIFAR-10 training, we use SGD with momentum of 0.9, an initial learning rate of 0.01, a batch size of 100, and 200 epochs. We use level smoothing of 0.1. We use the cosine annealing method to gradually decrease the learning rate while training. For CIFAR-100 training, we use a batch size of 128 and divide the learning rate by 5 at epochs \{50, 100, 130, 160\}. Other hyper-parameters are the same as those used in CIFAR-10. For Tiny ImageNet, we use a batch size of 256 with other hyper-parameters set to those used in CIFAR-100.

Training is performed using the HE-aware loss function in \cref{loss}. 
After completion of training, we prune for 100 iterations with fine-tuning. We set the initial threshold for both the position pruning and diagonal pruning to 0, then we gradually increase the threshold. Every iteration of pruning consists of three steps: (1) increasing the threshold; (2) removing groups of weights less than the threshold; and (3) fine-tuning. For fine-tuning, the learning rate is set to 1e-4. We do not utilize Lasso regularization at this stage. 


\subsection{Hyperparameter Exploration}



\begin{table}[t!]
\caption{Regularization Factor Search Space}
\begin{center}
\begin{sc}
\resizebox{\columnwidth}{!}{
\begin{tabular}{lccc}
Data Set     & $\lambda$      & $\lambda_p$               & $\lambda_d$                    \\ \hline
CIFAR-10     & \{0, 1, 5\}e-4 & \{0, 0.5, 1, 2, 4, 8\}e-4 & \{0, 5, 10, 15, 17.5, 20\}e-4 \\
CIFAR-100    & \{0, 1, 5\}e-4 & -                         & \{0, 1, 2, 5, 10, 20\}e-4  \\
Tiny ImageNet & \{0, 1, 5\}e-4 & -                         & \{0, 1, 2, 5, 10, 20\}e-4 
\end{tabular}
}
\end{sc}
\end{center}
\vskip -0.2in
\label{reg factor search space table}
\end{table}

We first study the effectiveness of Artemis on a variety of hyper-parameters as a way to understand the relative importance of position-based and diagonal pruning. Three hyper-parameter (regularization factors) are being studied: $\lambda$, $\lambda_p$, and $\lambda_d$. 
The search space used in our experiments is listed in \cref{reg factor search space table}. 

\cref{scatter} presents a full set of experiments for all hyperparameter values. It plots obtained model accuracy vs. the  number of rotations for all experiments on CIFAR-10. The x-axis shows the reduction in rotation number normalized to the full model (no pruning). In addition to individual experiment results (discrete points), we produce a curve that represents the Pareto frontier in the accuracy-cost space. \cref{cifar-10 full} also shows the best result of the baseline (Hunter) \footnote{While reproducing the results of \cite{hunter}, we discovered that the model can be pruned even further than reported}. 
While training, we noticed that using general (unstructured) regularization helps increasing accuracy of baseline models. Therefore, for fair comparison, we exclude the results having comparably higher accuracy at the beginning of the fine-tuning.)

Results in \cref{cifar-10 full} show that Artemis significantly increases potential for pruning. 
We define the primary region of interest to be solutions which suffer no more than 3\% accuracy degradation compared to the best accuracy.  In this region, Artemis is effective in reducing the number of $\Rotation$ operations by up to 3.5 times than the current state-of-the-art, at the same level of accuracy.

We also see that 
more aggressive regularization (using higher $\lambda_{p}$ and $\lambda_{d}$) results in lower accuracy at the end of training compared to the baseline. However, during the pruning-with-finetuning phase, they exhibit greater resilience to pruning.


We also empirically investigate the contributions of two rotation reductions strategies, position-based and diagonal pruning, on the optimal curves (the Pareto frontier).
\cref{cifar-10 simple} shows Pareto curves produced by the full Artemis and two modified versions in which only positional pruning or only diagonal pruning is deployed. (All curves are based on the same Lasso regularized training with only the pruning/fine-tuning step being different.) We see that Artemis with only diagonal pruning is close to Artemis with both pruning method while Artemis with positional pruning does not show any rotation number reduction. That permits us to conclude that diagonal pruning is the dominant factor in reducing the rotation number, and that position pruning does not contribute meaningfully to reducing the total cost. 

The dependence of rotation number on network parameters in \cref{cost} provides further insight into this result. If we take a typical structure of a CNN, such as, ResNet18, and consider the typical numbers of channels and the kernel size,  we find, for later layers of ResNet18, that $c_{out} \gg f$. Consequently, \cref{cost} shows 
shows that for similar values of $\alpha$ and $\beta$, the cost reduction is dominated by diagonal pruning. 


This finding is important as a way to reduce the cost of hyperparameter search involved in Artemis. If only diagonal pruning is sufficient for achieving cost reduction, there is no need to perform regularization to encourage position pruning. This eliminates the need to search over the values $\lambda_p$ during training.




\subsection{HE-Aware Model Compressability}



\begin{table}[t!]
\caption{Rotation count reduction (in percent): Hunter $\vert$ Artemis }
\begin{tabular}{cccc}
Models   & CIFAR-10 & CIFAR-100 & Tiny ImageNet \\ \hline
ResNet18 & 65.3 $\vert$ \textbf{90.2} & 11.2 $\vert$ \textbf{62.8} & 11.3 $\vert$ \textbf{25.6}              \\
ResNet50 & 42.5 $\vert$ \textbf{90.9} & 10.2 $\vert$ \textbf{55.4} &  20.4 $\vert$ \textbf{77.4}              
\end{tabular}
\label{trade-off table}
\end{table}

\cref{cifar-100} and \cref{tiny} show results for pruning of ResNet18 on two other datasets, CIFAR-100 and Tiny ImageNet. 
(Building on the above finding that position pruning is ineffective, we do not utilize position pruning and exclude $\lambda_p$ from the search space in all subsequent results.)
If, as before, we limit our interest to a region of 3\% accuracy degradation, we see in \cref{cifar-100} shows that Artemis is effective in reducing the number of $\Rotation$ operations by up to 2.4 times compared to Hunter. For Tiny ImageNet, \cref{tiny} shows Artemis being up to 1.2 times more effective in reducing $\Rotation$ operations.

However, what is clearly different in \cref{larger dataset resnet18} is that the results of both Artemis and Hunter suffer an accuracy loss at low pruning rates (small rotation number reduction). Intriguingly, this poor pruning rate of the model under HE-aware structure does not happen in plaintext pruning. \cref{resnet18 channel prune} shows the compressibility of ResNet18 on the Tiny ImageNet dataset. This indicates 
a genuine difference in the difficulty of pruning in plaintext and HE settings. 

We also use Artemis to reduce the number of rotation counts in a larger version of the ResNet model - ResNet50. 
In accordance with expectation, ResNet50, being larger, is easier to reduce the number of rotation counts even in the more restrictive HE setting, \cref{cifar-10resnet50}. Artemis can reduce the number of $\Rotation$ operations up to 6.3 times compared to Hunter, at same accuracy. \cref{cifar-100resnet50} and \cref{tinyresnet50} show that Artemis can reduce the number of $\Rotation$ operations up to 2 times and 3.5 times, respectively. 

The entire set of experimental results is summarized in \cref{trade-off table}, comparing the cost reduction produced by Hunter and Artemis. 
On a smaller model (ResNet18), Artemis does well on CIFAR-10 and CIFAR-100  but less so on Tiny ImageNet. On a larger model (ResNet50), Artemis succeeds in all datasets. In all cases, Artemis shows better results than Hunter. 

\section{Conclusion}

We proposed Artemis, a highly effective DNN pruning technique for HE-based inference. We judiciously investigated two HE-aware pruning strategies (positional and diagonal) to reduce the number of $\Rotation$ operations in HE convolution. We found that Pareto-optimal solutions are based fully on diagonal pruning. Artemis' benefits come from coupling DNN training, driven by a novel group Lasso regularization, with pruning to maximize HE-specific cost reduction (dominated by the $\Rotation$ operations). 
When evaluated on modern computer vision convolution models (ResNet18 and ResNet50) and several datasets, Artemis results in networks that reduce the number of $\Rotation$ operations by a factor of 1.2-6X higher compared to the state of the art.


\bibliography{main}
\bibliographystyle{icml2022}




\end{document}